# Autonomous Mobile Robot Navigation: Tracking problem

Husan F. Vokhidov
School of Science, Engineering and Environment
University of Salford
Manchester, United Kindom.
h.vokhidov@edu.salford.ac.uk

Salem Ameen
School of Science, Engineering and Environment
University of Salford
Manchester, United Kindom
s.a.ameen1@salford.ac.uk

***Abstract* - This study presents a study on autonomous robot navigation, focusing on three key behaviors: Odometry, Target Tracking, and Obstacle Avoidance. Each behavior is described in detail, along with experimental setups for simulated and real-world environments. Odometry utilizes wheel encoder data for precise navigation along predefined paths, validated through experiments with a Pioneer robot. Target Tracking employs vision-based techniques for pursuing designated targets while avoiding obstacles, demonstrated on the same platform. Obstacle Avoidance utilizes ultrasonic sensors to navigate cluttered environments safely, validated in both simulated and real-world scenarios. Additionally, the paper extends the project to include an Elegoo robot car, leveraging its features for enhanced experimentation. Through advanced algorithms and experimental validations, this study provides insights into developing robust navigation systems for autonomous robots.**

## I. Introduction

In the realm of robotics, the pursuit of innovation and advancement drives researchers to explore novel techniques and methodologies to enhance the capabilities of robotics systems. This investigation delves into the integration of advanced navigation and control methodologies in the development of a mobile robot platform, aiming to push the boundaries of autonomous and versatility in robotics applications.

The motivation for this investigation stems from the very-growing demand for sophisticated robotics systems capable of navigating complex environments and performing diverse tasks with precision and efficiency. As technology continues to evolve, there is a pressing need to explore new avenues for enhancing the functionality and adaptability of robots to meet the demands of various industries and domains.

The primary objective of this paper is two twofold: first, to explore the feasibility and effectiveness of integrating advanced navigation techniques such as Odometry, Object Tracking, and Obstacle Avoidance into a mobile robot platform; and second, to develop a user control program that enables seamless interaction and communication with the robot view Wi-Fi. By achieving these objectives, the paper aims to demonstrate the potential of these advancements in revolutionizing robotics applications across multiple domains.

The significance this investigation lies in its potential to address real-world challenges and catalyze advancements in robotics technology. By equipping mobile robot platforms with advanced navigation and control capabilities, researchers and practitioners can unlock new possibilities in fields such as autonomous transportation, industrial automation, search and rescue operations, and environmental monitoring. Furthermore, the development of a user control program facilitates remote operation and data exchange, enhancing the usability and accessibility of robotic systems in various settings.

The contributions of this research extend beyond the confines of academia, offering practical solution and insights that can be leveraged by educational institutions, industries, governments, and research institutions alike. By harnessing the power of navigation and control techniques, the mobile robot platform developed in this study holds the capability to enhance existing paradigms and pave the way for a future where robots play a pivotal role in shaping our society and advancing human progress.

### A. Literature Review

The field of robotics has witnessed rapid advancement in the past few years, driven by the convergence of cutting-edge technologies and innovative research efforts. In this literature review, we explore fundamental investigations and developments that have shaped the landscape of mobile robotics, focusing on advanced navigation and control techniques.

Odometry has long been a cornerstone of mobile robot navigation, providing essential feedback for estimating position and orientation based on wheel encoder measurements. Research by Thrun et al. (2005) [1] demonstrated the efficacy of probabilistic techniques such as particle filters in addressing odometric uncertainties, paving the way for more robust localization algorithms in mobile robots.

Object tracking has emerged as a critical capability for mobile robots operating in dynamic environments. Recent studies by Milan et al. (2016) [2] have explored the application of deep learning techniques, specifically convolution neural networks (CNNs), for real-time object detection and tracking. By leveraging CNNs trained on large-scale datasets, researchers have achieved remarkable accuracy in identifying and tracking objects of interest, enabling robots to navigate complex scenarios with greater efficiency and reliability.

The integration of Wi-Fi communication capabilities into mobile robot platforms has enabled new opportunities for remote operation and data exchange. Studies by Liu et al. (2018) [3] have explored the use of Wi-Fi based localization techniques, leveraging signal strength measurements to estimate the robot's position in indoor environments with minimal infrastructure requirements. Additional, advancements in cloud robotics have enabled seamless integration with could-based services for enhanced perception, planning, and decision-making capabilities (Kadous et al., 2019) [4].

One fundamental aspect of mobile robotics is localization, which involves determining the robot's position and orientation relative to its surroundings. Traditional approaches such as odometry and inertial navigation have been complemented by more sophisticated techniques like Simultaneous Localization and Mapping (SLAM). Notable studies by Dissanayake et al. (2001) [5] and Thrun et al. (2005) [6] have demonstrated the efficacy of probabilistic SLAM algorithms in enabling robots to autonomously map and navigate unknown environment.

Object detection and tracking are now considered crucial functions for mobile robots navigating complex surroundings. Recent strides in computer vision, notably the widespread adoption of deep learning methods, have transformed how robots recognize objects. Research conducted by Redmon et al. (2016) [7] and Liu et al. (2018) [8] has demonstrated the effectiveness of convolutional neural networks (CNNs) for detecting objects in real time, empowering robots to identify and track objects swiftly and accurately.

Furthermore, the integration of communication features, such as Wi-Fi and cloud-based services, has greatly enhanced the capabilities of mobile robot platforms. Investigations by Gao et al. (2017) [9] and Kadous et al. (2019) [10] have delved into the application of Wi-Fi based localization and could robotics to improve perception, planning, and decision-making processes. By utilizing cloud resources, robots can delegate computationally intensive tasks and access vast datasets, leading to better situational awareness and decision-making capabilities.

To sum up, advancements in navigation, control, and communication have propelled mobile robotics to unprecedented levels of performance, enabling robots to operate autonomously in various environments with remarkable accuracy and efficiency. The incorporation of probabilistic localization, deep learning-based object detection, and cloud-based communication sets the stage for mobile robots to take on central roles in applications such as autonomous transportation, surveillance, and environmental monitoring.

### B. *Paper Outline*

The remainder of the paper is structured as follows:

Section II provides a comprehensive model description, detailing the Odometry, Target Tracking, and Obstacle Avoidance methods utilizing a Pioneer robot and Aria simulator. In Section III, our Contribution Work is presented, encompassing the implementation of the same methods on another smart robot car. This section also discusses extended investigations as our unique contribution for this paper.

Section IV presents the Experimental Results and Analysis, showcasing the performance evaluation of both our robots. Finally, Section V offers Conclusions and Future Directions, summarizing the findings and outlining potential avenues for future research.

## II. METHODS

The principal objective of this endeavour is to conceptualize, program, and evaluate three distinct robot behaviours aimed at autonomously guiding the robot through a predefined environment to a designated destination. The subsequent subsections delineate the three behaviours slated for implementation.

### A. *Odometry*

#### 1) *Method description*

The odometry model serves as a fundamental component within the realm of mobile robotics, particularly facilitating navigation for robots equipped with differential drive configuration. Operating within a framework of vector coordinate pairs, commonly referred to as nodes or waypoints, this model enables precise traversal of predefined paths.

The navigation method capitalizes on an odometer/trajectory model, wherein the robot's movement is meticulously orchestrated based on the acquired odometric data. The equations for odometry typically calculate changes in position and orientation based on wheel encoder readings. The basic equations can be written as shown Equ. 1, as described in [11].

$$\Delta d = \frac{1}{2}(d_{\text{left}} + d_{\text{right}}) \tag{1}$$

Equation 1: Incremental Distance Travelled

Where $d_{left}$ and $d_{right}$ are the distance travelled by the left and right wheels, respectively.

Its functionality is the acquisition of odometric data encompassing the robot's positional coordinates (x, y) and orientation angle. Leveraging information from wheel encoder sensors and the intrinsic geometry of the robot, this data is harnessed to compute crucial navigation parameters. Specifically, the model employs a two-fold approach: first, it determines the angular transition necessary to orient the robot towards the target node, a calculation often predicated on the arc-tangent function. The basic equation can be calculated as shown in Equ. 2 as described in [12].

$$\Delta\theta = \frac{d_{\text{right}} - d_{\text{left}}}{\text{wheelbase}} \tag{2}$$

Equation 2: Change in orientation.

Where the *wheelbase* is the distance between the two drive wheels.

Subsequently, a linear transition, guided by the Euclidean distance between nodes, propels the robot towards its intended destination. This methodology not only underscores the intricate interplay between sensor data and geometric principles but also underscores the robustness of odometry-based navigation techniques in guiding mobile robots along predefined trajectories.

Changing X and Y coordinates and assigning new X and Y coordinates can be done as described in [13] and [14]. The equations are shown as follows:

$$\Delta x = \Delta d \times \cos(\theta + \frac{\Delta\theta}{2})$$
$$\Delta y = \Delta d \times \sin(\theta + \frac{\Delta\theta}{2}) \quad (3)$$

Equation 3: Change in X and Y coordinates.

$$x_{\text{new}} = x_{\text{old}} + \Delta x$$
$$y_{\text{new}} = y_{\text{old}} + \Delta y \quad (4)$$

Equation 4: New X and Y coordinates.

These equations provide a basis for estimating the robot's position and orientation using odometry data from wheel encoders. They are fundamental to odometry-based navigation systems and are commonly used in robotics applications.

Central to this methodology is the concept of robot velocity, a dynamic parameter that dictates the velocity and orientation of the robot's motion. Through careful manipulation of velocity, synchronized with angular and linear transitions, the robot seamlessly traverses from one waypoint to another.

Moreover, our implementation program incorporates a mechanism to assess navigation complexity.

### 2) *Experimental setup*

In our inaugural experiment on robot navigation architecture, we employ a Pioneer robot using java coding environment as stipulated by the paper guidelines.

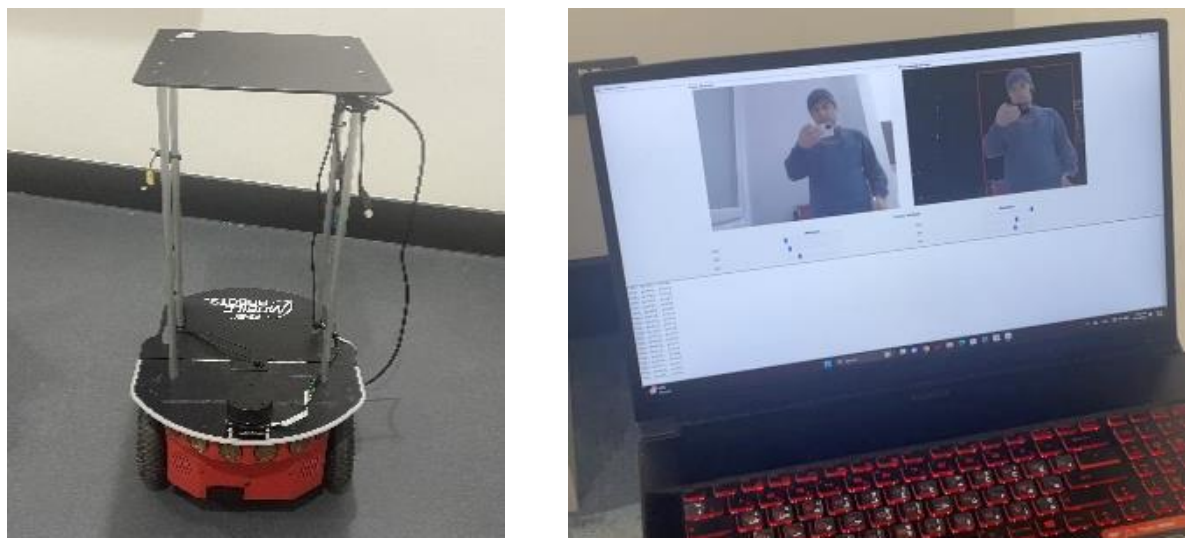

Figure 1. Pioneer Robot(a) and Control Panel Interface (b).

Experimental trials were conducted to assess the efficacy of the odometry-based navigation methodology using both simulated and real-world environments. Initially, experiments were conducted using the Aria simulator to validate the functionality of the navigation algorithms in a controlled virtual environment. Subsequently, trials were performed with a Pioner robot to assess the methodology's efficacy in real-world scenarios.

The Aria simulator provided a platform for preliminary experimentation before deploying the methodology on the Pioneer robot. MobileSim Simulator map, featuring an environment size of 8.95mx4.65m, was utilized for conducting the simulated trials. The environment comprised a room layout with distinct features, including an inner room with a king-size bed and an entrance door leading to a kitchen via a narrow and long passage. Notably, the entrance door remained open throughout the experiments to simulate real-world conditions accurately.

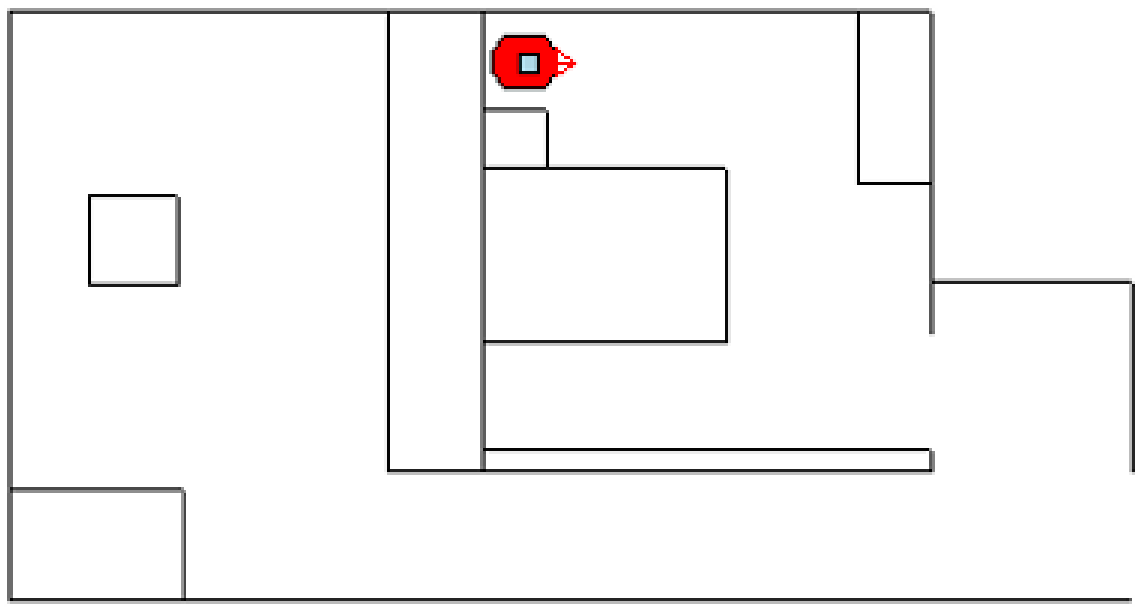

Figure 2. MobileSim Simulator Map (Source: Salford University Mobile Robotics course material).

Following successful validation in the simulation environment, the methodology was deployed on a Pioneer robot for real-world experiments. The experiments were conducted in a room with similar dimensions to the simulated environment, providing consistency in the testing conditions, The real-world environment mirrored the layout of the simulated environment, including the presence of obstacles and the configuration of layout. Additionally, the environment was equipped with a variety of obstacles strategically placed to challenging navigation scenarios.

In our implementation, we augmented the odometry model with waypoints to further refine the navigation process. These waypoints, represented as coordinates in the form of Node = {x, y}, such as {2200, 0}, {2268, -2387}, {4315, -2387}, {4315, -3649}, {-1285, -3873}, serve as pivotal landmarks guiding the robot along its designated path.

During both simulated and real-world experiments, data collection focused on capturing relevant metrics every 100ms such as robot coordinates, orientation, right-left wheel velocities etc. The gathered data served as a foundation for evaluating the effectiveness of the odometry performance. The gathered data served as a foundation for assessing the effectiveness of the odometry-based navigation methodology in both simulated and real-world environments.

## B. *Target Tracking*

In this subsection, we outline the methodology employed for tracking, a pivotal component of autonomous robot navigation. Target tacking involves the robot's ability to detect, follow, and utilizing camera sensors, the robot identifies a colour blob selected by the user as the target for tracking. The subsequent behaviour encompasses three discrete zones within the image frame: identification, pursuit, and approach. Throughout these zones, the robot dynamically adjusts its velocity to manoeuvre towards the target while avoiding obstacles. The following sections detail the implementation and experimental setup for target tracking.

#### 1) *Method description*

The target tracking algorithm is encapsulated within the tracking method, which orchestrates the robot's behaviour based on the detected target's position within the image frame. The method receives a velocity parameter to regulate the robot's speed during tracking.

The robot divides the image frame into multiple segments and identifies the target's position within these segments. If the target is in the left segment, the robot initiates a right turn towards the target. Conversely, if the target is in the right segment a left turn is executed.

In this zone, the robot dynamically adjusts its trajectory to align with the target's movement. By continuously monitoring the target's position, the robot manoeuvres towards the target while maintaining a constant velocity.

As the robot approaches the target, it regulates its velocity to ensure a safe distance is maintained. Upon reaching a predefined threshold distance from the target, the robot halts its motion, indicating the target has been successfully approached.

#### 2) *Experimental setup*

The target tracking trials were carried out employing a physical Pioneer robot outfitted with camera sensors. The experiments were conducted in a room measuring 3.00mx4.68m, containing various obstacles and a designated target area.

#### 3) *Experimental setup*

The provided code snippet demonstrates the integration of the target tracking algorithm within the robot's control framework. The *track()* method is invoked within the main control loop after finishing odometry model, where the robot continuously monitors the target's position and adjusts its velocity accordingly. Additionally, the experiment's data logging functionality records the robot's odometry data and target racking performance for subsequent analysis.

The target tracking methodology outlined herein provides a robust framework for autonomous robot navigation, enabling efficient detection, pursuit, and approach of designated targets within diverse environments. Trough vision-based technology and dynamic velocity control, the robot demonstrates adept target racking capabilities, essential for real-world applications in surveillance, exploration, and human-robot interaction.

### C. *Obstacle Avoidance*

The obstacle avoidance behaviour plays a critical role enabling autonomous robots to navigate effectively in dynamic environments. This method equips the robot with the capability to detect and circumvent various obstacles encountered during its traversal, including furniture, walls, and other objects. Leveraging a sonar ring comprising eight ultrasonic range sensors, the avoidance algorithm facilitates real-time obstacle detection and evasion. This methodology encompasses three primary functionalities: avoidance, decollation, and entrapment escape, each aimed at ensuring the robot's safe navigation amidst obstacles.

#### 1) *Method description*

The method collects range measurements from the left and right ultrasonic sensors to assess the proximity of obstacles. By comparing these measurements against a predefined threshold, the algorithm identifies potential obstacles in the robot's path.

A corner trapping scenario is detected when obstacles are simultaneously detected on both the right and left sides. In such cases, the robot executes an escape manoeuvre, involving a backward motion followed by a random turn to disengage from the corner trap.

If obstacles are detected either on the left or right side, the robot initiates a corresponding turn away from the obstacle to avoid collision. Additionally, special provisions are made to handle scenarios where the robot becomes stuck due to obstacles or navigation constraints. In such cases, the robot executes an escape manoeuvre to resume its movement.

#### 2) *Experimental Setup*

The obstacle avoidance experiments were conducted using both simulated environments in the Aria simulator and real-world scenarios with a physical Pioneer robot. The experiments were conducted in a room containing various obstacles representing of typical indoor environments. The performance of the avoidance algorithm was evaluated under different obstacle configurations to assess its effectiveness in navigating complex environments.

The obstacle avoidance methodology presented offers a robust framework for enabling autonomous robots to navigate safely in cluttered environments. By leveraging sonar-based sensing and dynamic manoeuvre techniques, the algorithm facilitates real-time detection and evasion of obstacles, thereby ensuring the robot's smooth traversal towards its intended destination. These capabilities are instrumental in enhancing the autonomy and adaptability of robotic systems in diverse real-world scenarios.

## III. Contribution

### A) *Experiment robot.*

To extend our project, we utilized a robot car manufactured by Elegoo. This versatile robot is equipped with various features, making it an ideal platform for experimentation and learning. The key components of the robot include:

#### 1) *ELEGOO Smart Robot Car Kit V4.0*

This kit is compatible with Arduino IDE and comes with an UNO R3 board as its core controller. It provides a solid foundation for building and programming the robot.

#### 2) *Ultrasonic Sensor*

The robot is outfitted with an ultrasonic sensor for precise distance measurement. This sensor enables the robot to detect obstacles in its path and navigate effectively.

*3) Ultrasonic Sensor*

With the line tracking module, the robot can follow predefined path or lines on the ground. This feature is particularly useful for tasks such as maze-solving and line-following challenges.

*4) IR Module*

The IR module enhances the robot's ability to interact with its environment by detecting infrared signals. This enables additional functionalities such as remote control or communication with other IR-enabled devices.

*5) DC Motors and Motor Driver*

The robot is equipped with four wheels driven by DC motors. These motors can be controlled using a motor driver, allowing for precise movement and manoeuvrability.

*6) ESP32 Wi-Fi Module with Camera*

One of the standout features of the robot is ESP32Wi-Fi module coupled with a camera. This enables communication and steaming of video footage, opening possibilities for remote operation, surveillance, and image processing applications.

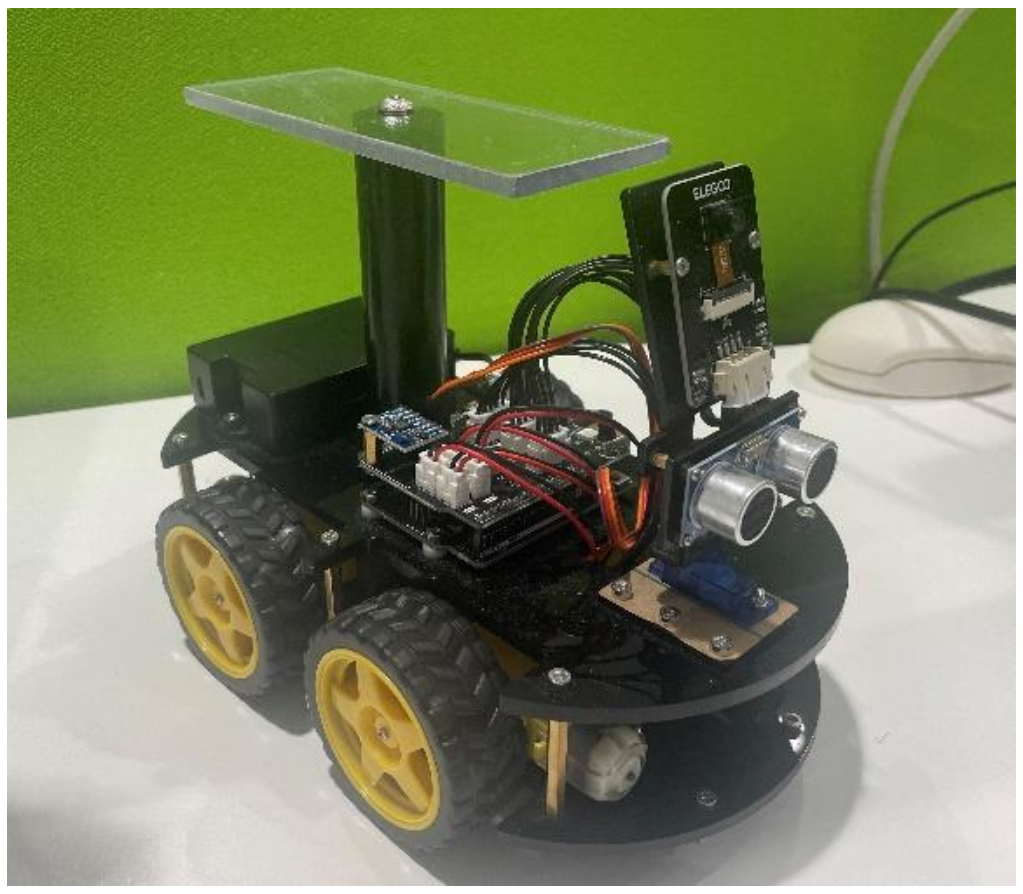

Figure 3: ELEGOO Smart Robot Car Kit V4.0 with Arduino IDE compatibility and key components.

Additionally, the remote communication protocol enables users to develop the main program in any kind of programming language such as our case Python and execute it on their PC or laptop, enhancing the versatility and accessibility of the robot car. The robot car offers a comprehensive platform for hands-on exploration robotics, electronics, and programming concepts. Its educational value makes it an excellent choice for students, hobbyists, and enthusiasts alike.

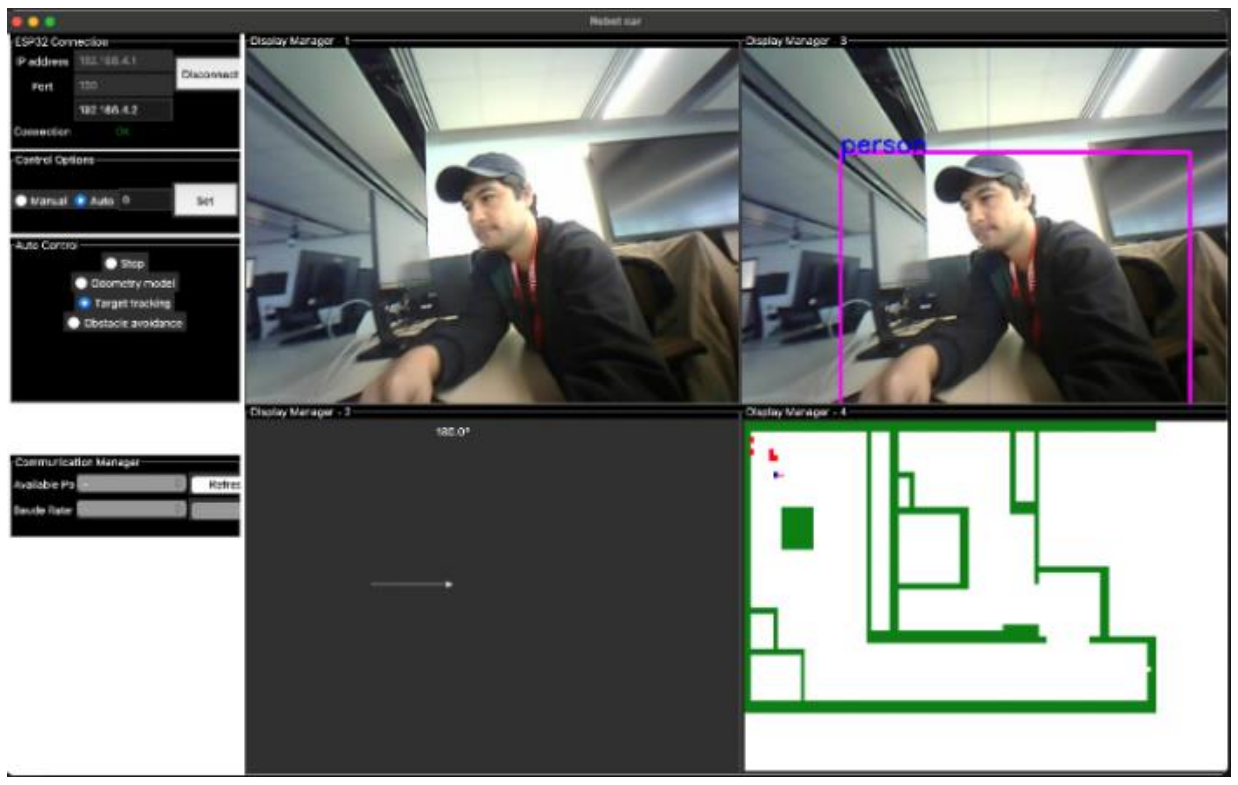

Figure 4: Control program interface overview.

*B) User control program.*

We developed a user interface program to facilitate the remote control of the robot and the reception of sensor data or video images from the robot car.

*1) The Connection*

The ESP32 Wi-Fi module is equipped with its built-in Wi-Fi module, enabling wireless communication with the control program running on a PC, laptop, or a smartphone. Pairing with a camera module, the Wi-Fi module transforms the robot into an IoT device capable of capturing and transmitting images or video over a Wi-Fi network. This combination allows for creation of a remotely controllable robot application integrated with image processing techniques. Connection can be established by joining the ESP32 vie Wi-Fi.

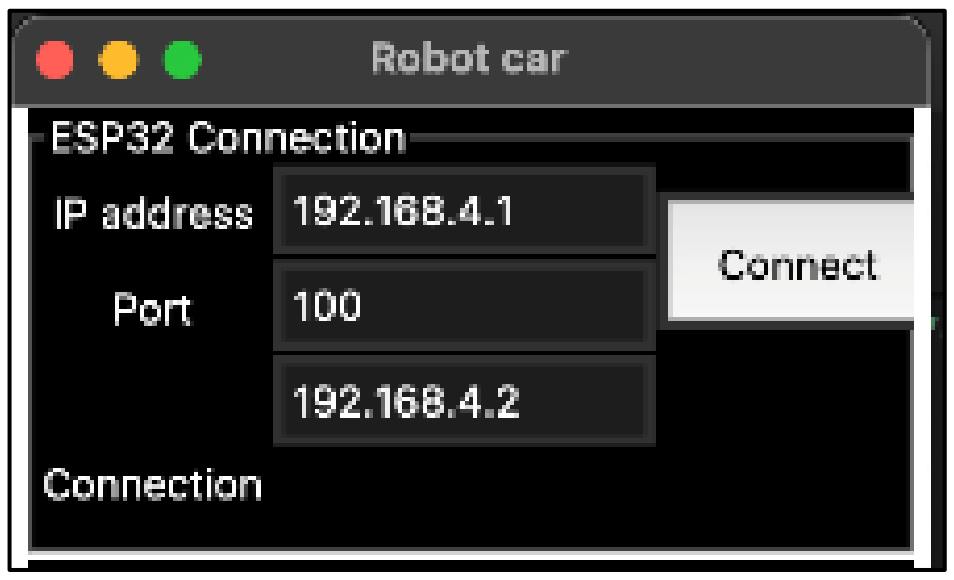

Figure 5: Wi-Fi connection panel.

*2) Communication and protocols*

The communication between different components of the robot is facilitated by specifically designed communication protocols implementation in C++ code and executed on an Arduino-compatible microcontroller board.

*3) Control Panel*

The control panel is designed to enable users to control the robot manually or set it to operate fully autonomously. By selecting the *"manual"* control option, users can manoeuvre the robot car in all directions or rotate the camera head from right to left.

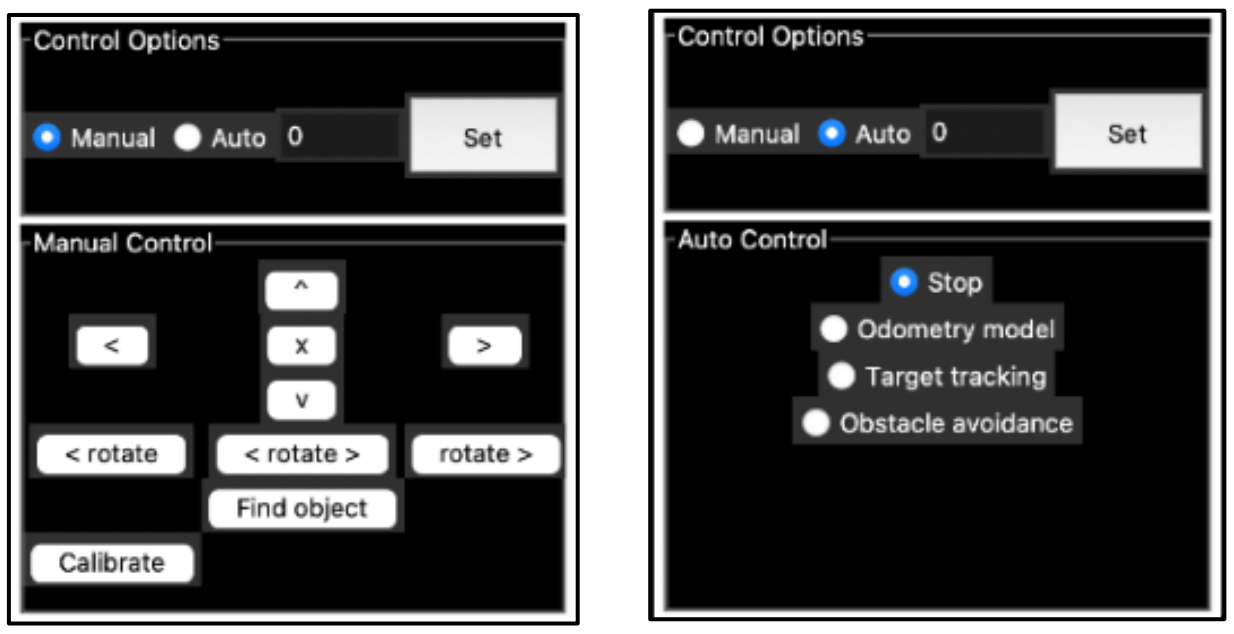

Figure 6: Control options for Manual (a) and Auto (b) control.

Additionally, users can detect and identify objects from camera images by just pressing a button on the control panel or a specific key on the keyboard. The autonomous option comprises *"Odometry"*, *"Target tracking"*, *"Obstacle avoidance"* functionalities.

## C) Odometry

Navigating through complex environments poses significant challenges for autonomous robots, necessitating efficient strategies for path planning and navigation. In our approach, we leveraged a combination of advanced algorithms and intuitive visualization to facilitate the seamless movements of robots within their surroundings.

### 1) Odometry algorithms

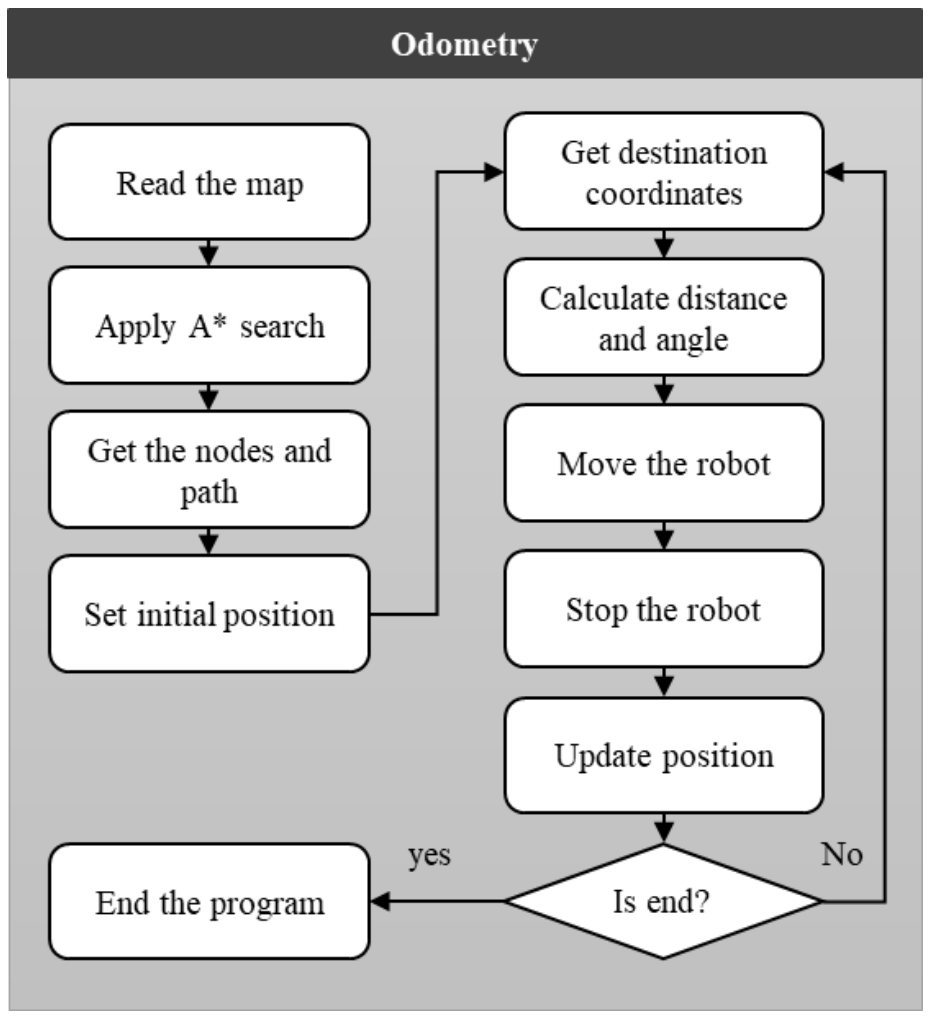

Figure 7: Flowchart illustration the odometry algorithms for autonomous robot navigation.

Central to our methodology is the representation of map information stored within a text file, as illustrated in Fig. (a). Within this file, characters such as "*#*", "*M*", and "*E*" hold significant meaning. "*#*" signifies the presence of walls or objects obstructing the robot's path, "M" designates the initial point of the robot's journey, and "E" marks the endpoints towards which the robot navigates.

In contrast to traditional manual methods of inputting odometry coordinates, we adopt A* search algorithm to efficiently discover the optimal route from the initial position to the endpoint, as portrayed in Fig. (b). Here, the path coordinates are visually represented by "*" characters, offering a clear depiction of the route generated by the search algorithm.

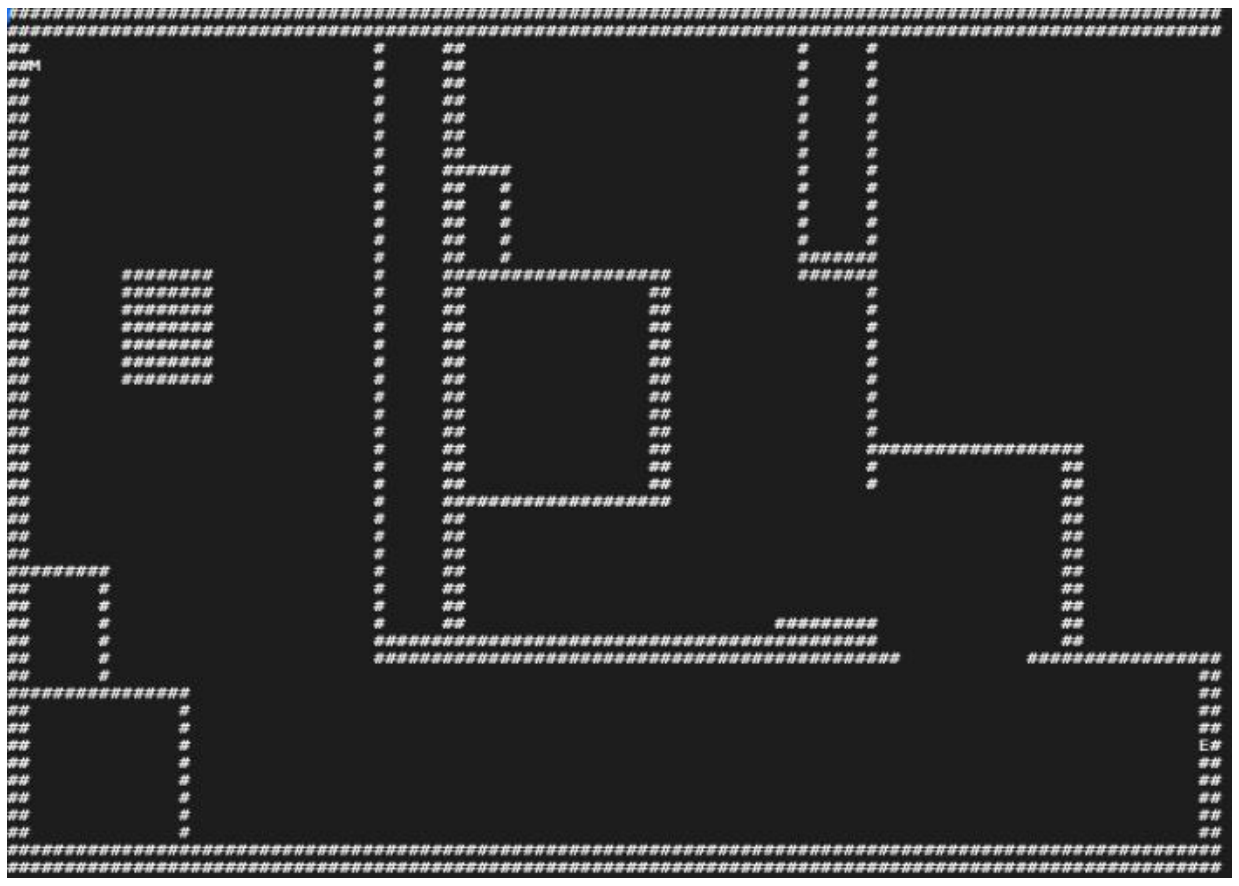

(a)

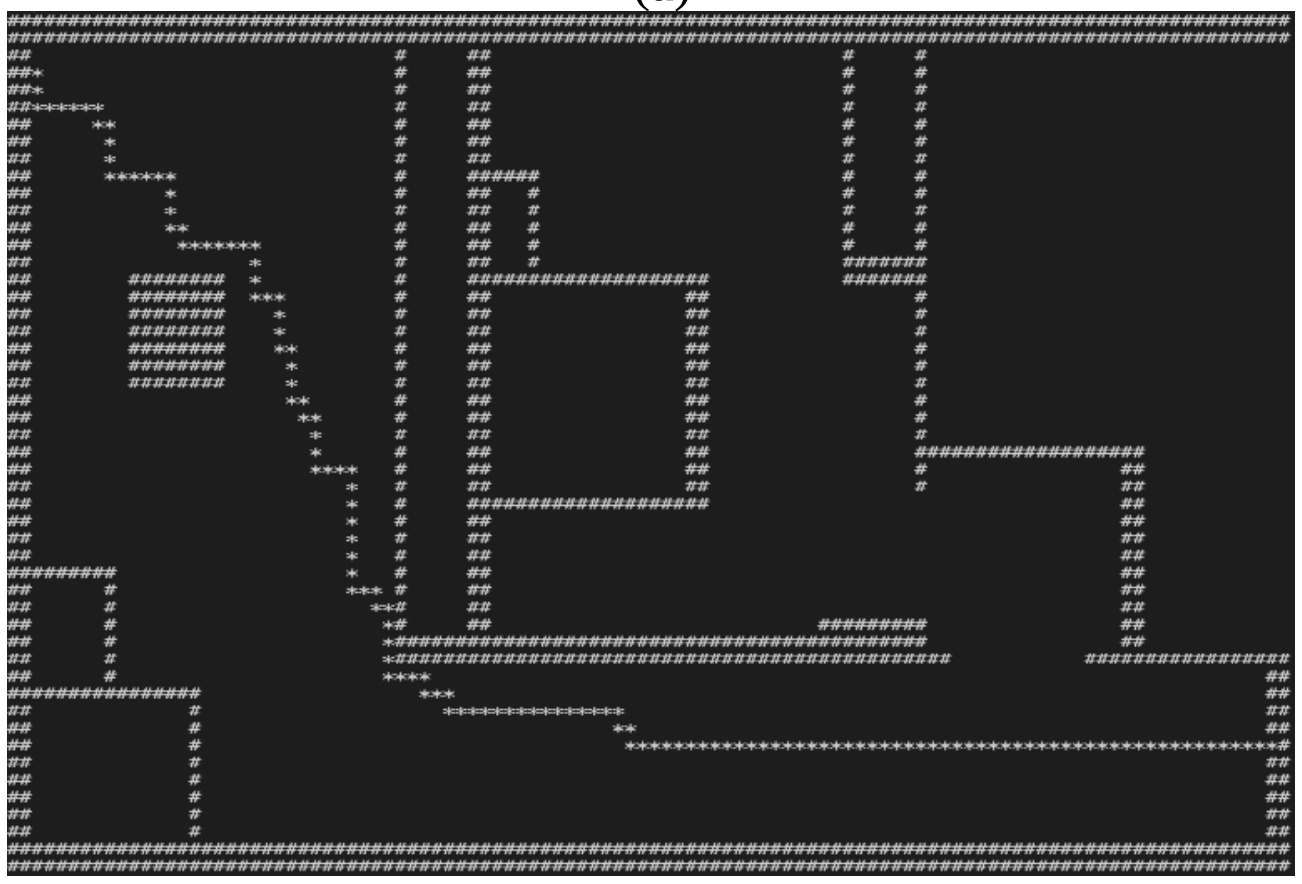

(b)

Figure 8. The view of the map in a text file (a) and the path created by A* algorithm (b).

It's important to note that Elegoo robot car, unlike Pioneer robot, does not come equipped with wheel encoders, introducing an additional layer of complexity to the odometry process. Additionally, the wheels of the Elegoo robot car my not provided precise movements, and variations in speed con occur depending on the battery power level.

Upon path generation, the robot transitions into autonomous operation, traversing along the path's nodes determined by the A* algorithm. At each juncture, the robot gracefully executes a 90-degree turn, ensuring precise navigation through the environment. To provide a tangible representation of the robot's movement along the predetermined path, a simulation map is crafted, showcased in Fig. _. The map employs red points to highlight pivotal turning points along the route, while a solitary blue point symbolizes the physical robot. Additionally, a slender red turret extends from the robot, indicating its current orientation.

Furthermore, to optimize the robot's trajectory, any extraneous nodes between turning nodes are strategically removed, streamlining the robot's overall movement, and enhancing its efficiency within the environment.

By combining advanced algorithms with intuitive visualizations, our approach offers a comprehensive solution for autonomous robot navigation, enabling smooth and efficient movement through complex environments.

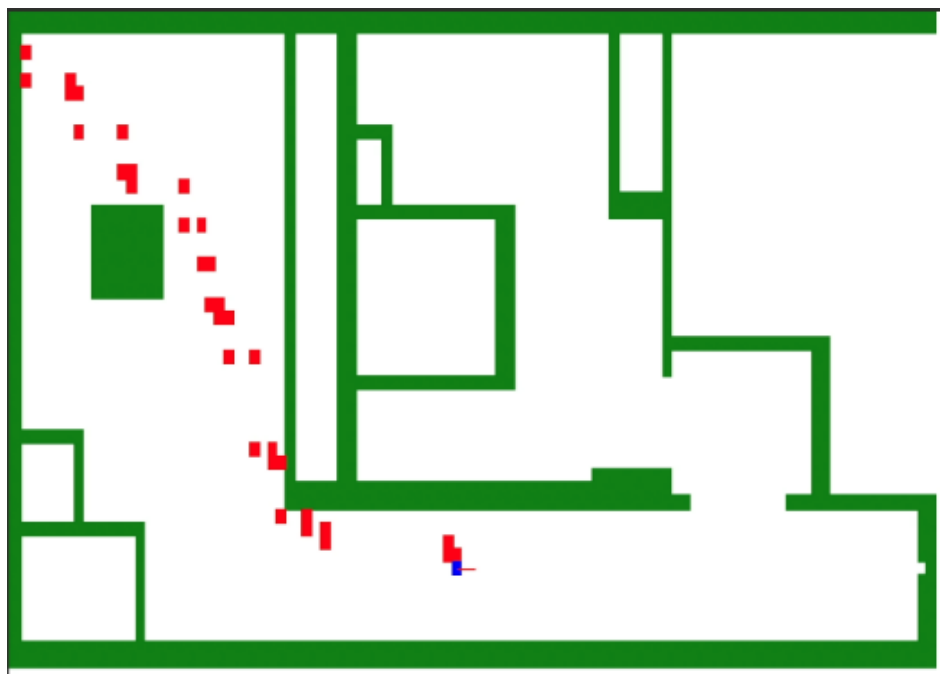

Figure 9. The simulation map. The red points are where robot make turns and the blue point is the robot.

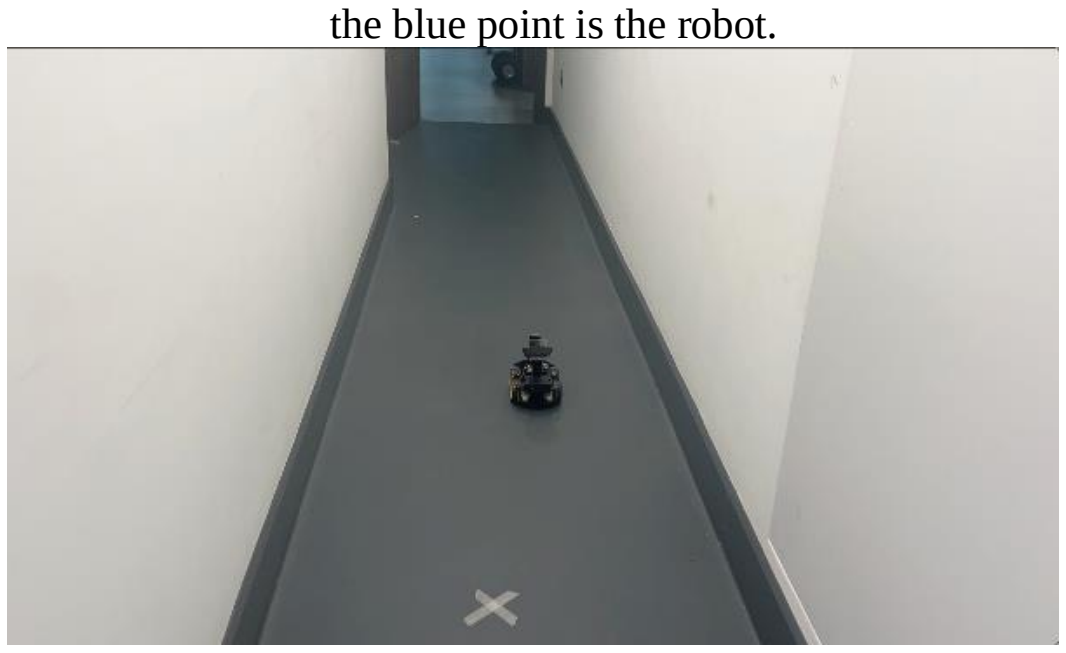

Figure 10: Example of an autonomous robot navigating through a passage

*D) Object Tracking.*

*1) Object tracking algorithm.*

The object tracking algorithm mainly consists of two steps: object searching and object tracking alongside obstacle avoidance.

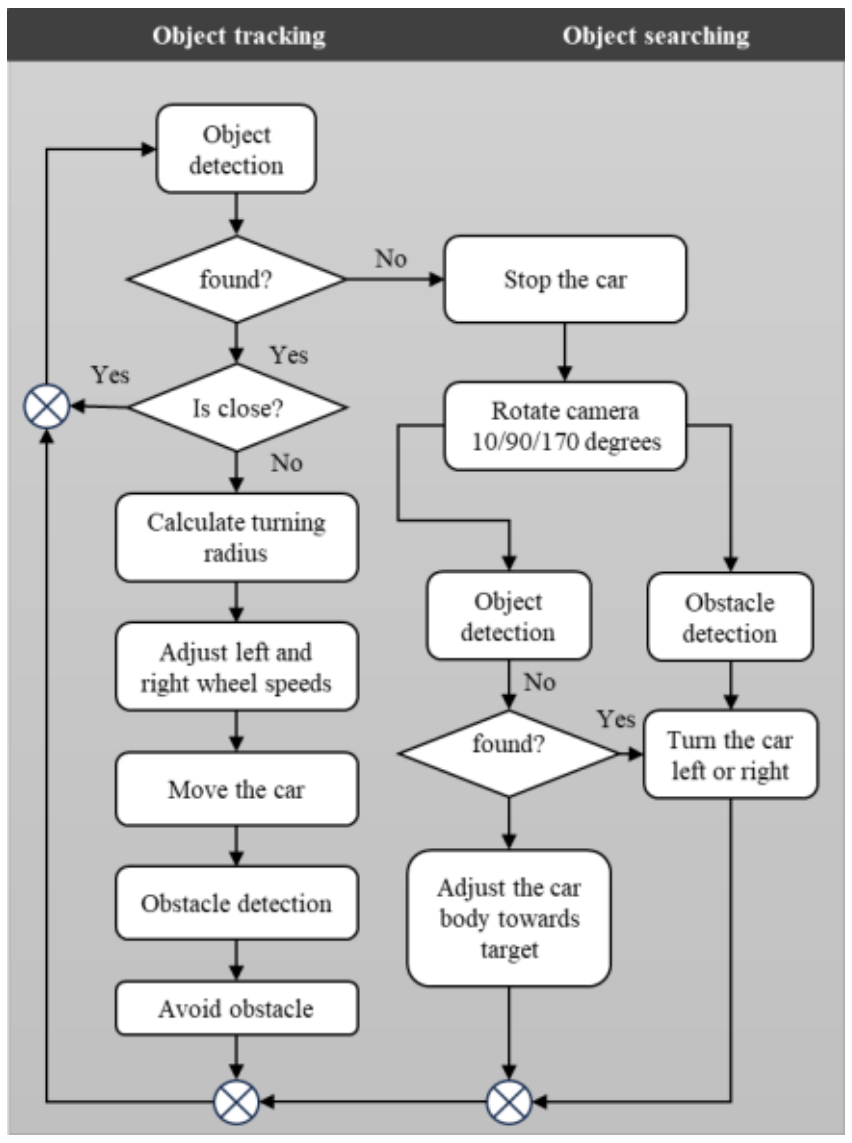

Figure 11: Flowchart illustration for the autonomous robot Object Tracking.

To execute the object detection task, the 'YOLOv8' object detection deep-learning algorithm is applied. This algorithm involves detecting and identifying a desired object in our project. To accomplish this, a stream image is captured and transmitted remotely over a Wi-Fi network to the control program, which runs the 'YOLOv8' object detection model on a PC or a laptop, prepared to execute the object detection task.

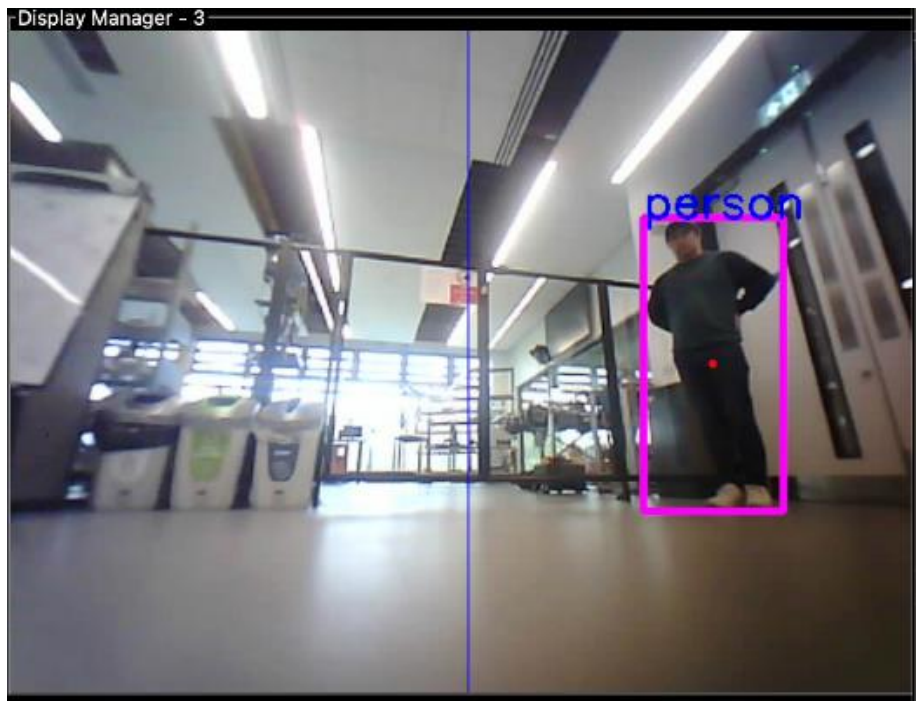

Figure 12: Example of an autonomous robot object detection

Upon execution of the object detection task, a set of bounding boxes of the desired object is obtained as a result, along with class labels and confidence scores. The target object could be anything based on the trained data. However, since we have utilized the pretrained model of YOLOv8 on the COCO dataset, 80 objects can be recognized as target objects.

To simplify computation, the algorithm is configured to choose only a single target object with the highest score. By default, the 'person' class is set as the default target object, which is also utilized in our demo application.

*E) Obstacle Avoidance*

To avoid obstacle for our robot car, an ultrasonic sensor data is used to measure distance to nearby objects. When an object is detected within a certain range, obstacle avoidance algorithm is applied to change robot car's direction.

*1) Object avoidance algorithm.*

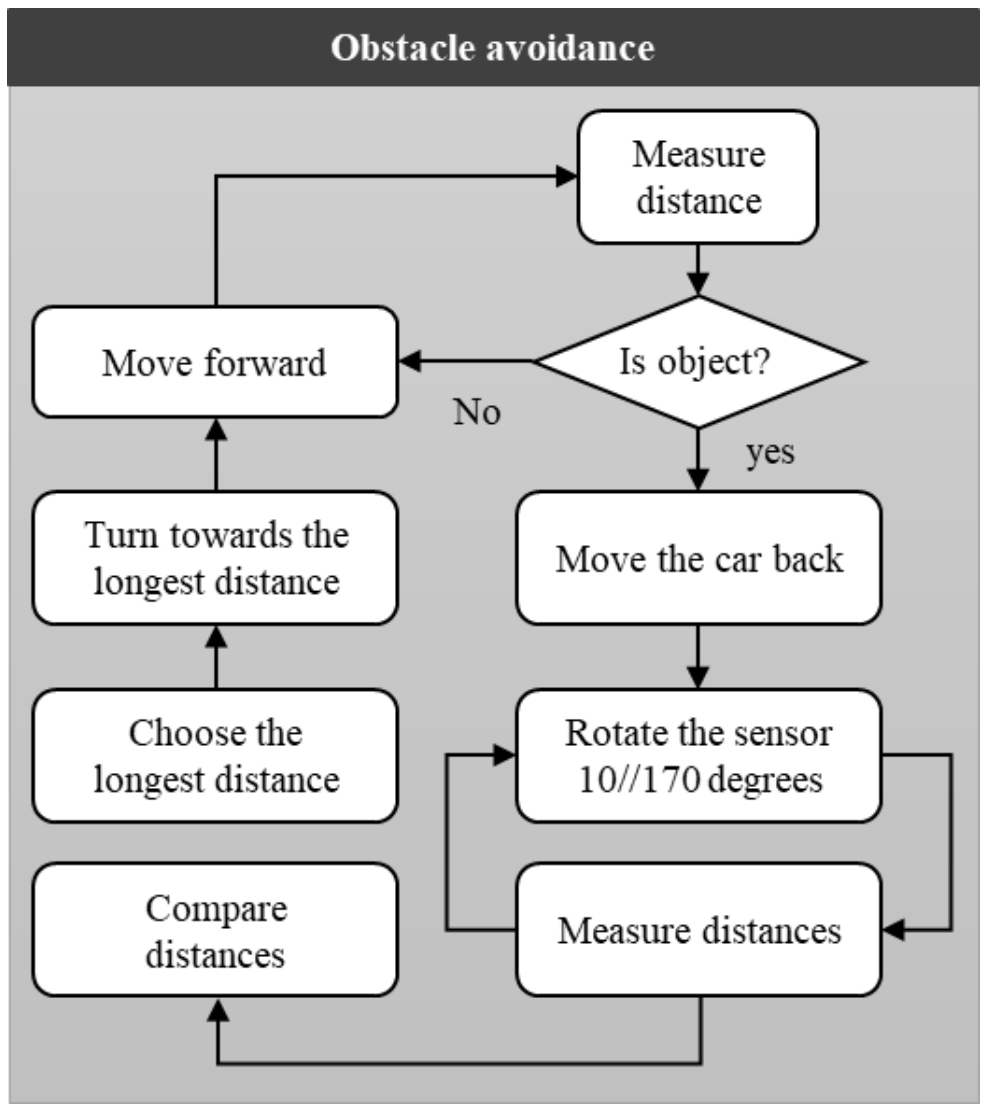

Figure 13: Flowchart illustration for the autonomous robot Object Avoidance.

The algorithm for obstacle avoidance is straightforward. Initially, the distance between the obstacle and the robot car is measured. If the distance falls below a certain range, the car is halted and moved back slightly. Subsequently, the distances among the robot car and barriers on the right and left sides are measured by rotating the ultrasonic sensor at angles of 10 and 70 degrees, respectively. These two distances are then compared to select the longest one. Finally, the car is turned towards the direction of the longest distance and proceeds forward. By iteratively repeating this process, the robot car can autonomously navigate without colliding with any obstacles.

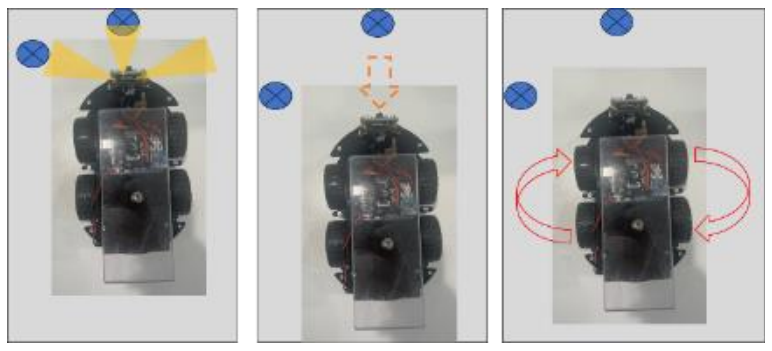

Figure 14: Flowchart illustration for the autonomous robot Object Tracking.

| N | X | Y | Th | l_vel | r_vel | nTh | Tracking |
|---|---|---|---|---|---|---|---|
| 1 | 0 | 0 | 0 | 0 | 0 | 0 | FALSE |
| 2 | 0 | 0 | 0 | 0 | 0 | 0 | FALSE |
| 3 | 0 | 0 | 359.9166 | 20 | 15 | 0 | FALSE |
| 4 | 3 | 0 | 359.9166 | 35 | 34 | 0 | FALSE |
| 5 | 10 | 0 | 359.9166 | 70 | 72 | 0 | FALSE |
| … | ... | … | … | … | … | … | … |
| 180 | 2119 | 0 | 0 | 114 | 115 | 273 | FALSE |
| 181 | 2128 | 0 | 359.9166 | 86 | 85 | 273 | FALSE |
| 182 | 2134 | 0 | 359.7408 | 66 | 43 | 273 | FALSE |
| … | ... | … | … | … | … | … | … |
| 439 | 2246 | -2309 | 272.9038 | 112 | 112 | 357 | FALSE |
| 440 | 2246 | -2318 | 272.9038 | 84 | 82 | 358 | FALSE |
| 441 | 2247 | -2323 | 273.2554 | 39 | 66 | 358 | FALSE |
| … | ... | … | … | … | … | … | … |
| 1356 | -1175 | -3834 | 182.8148 | 119 | 120 | 199 | FALSE |
| 1357 | -1186 | -3835 | 182.8148 | 120 | 120 | 200 | FALSE |
| 1358 | -1198 | -3836 | 182.8148 | 117 | 118 | 203 | FALSE |
| 1359 | -1210 | -3836 | 182.8148 | 121 | 121 | 0 | TRUE |
| 1360 | -1221 | -3837 | 182.8148 | 98 | 100 | 0 | TRUE |
| 1361 | -1228 | -3837 | 182.8148 | 72 | 77 | 0 | TRUE |
| … | ... | … | … | … | … | … | … |
| 1507 | -1607 | -3631 | 132.9802 | 39 | 40 | 0 | TRUE |
| 1508 | -1609 | -3628 | 132.9802 | 37 | 39 | 0 | TRUE |
| 1509 | -1612 | -3625 | 132.9802 | 40 | 39 | 0 | TRUE |
| … | ... | … | … | … | … | … | … |
| 2238 | -2823 | -1260 | 126.0367 | 41 | 39 | 0 | TRUE |
| 2239 | -2826 | -1257 | 126.0367 | 40 | 40 | 0 | TRUE |
| 2240 | -2828 | -1254 | 126.0367 | 38 | 37 | 0 | TRUE |

## IV. RESULTS

TABLE I
EXPERIMENTAL TABLE FOR THE FIRST EXPERIMENT FOR THE ODOMETRY AND THE TARGET TRACKING.

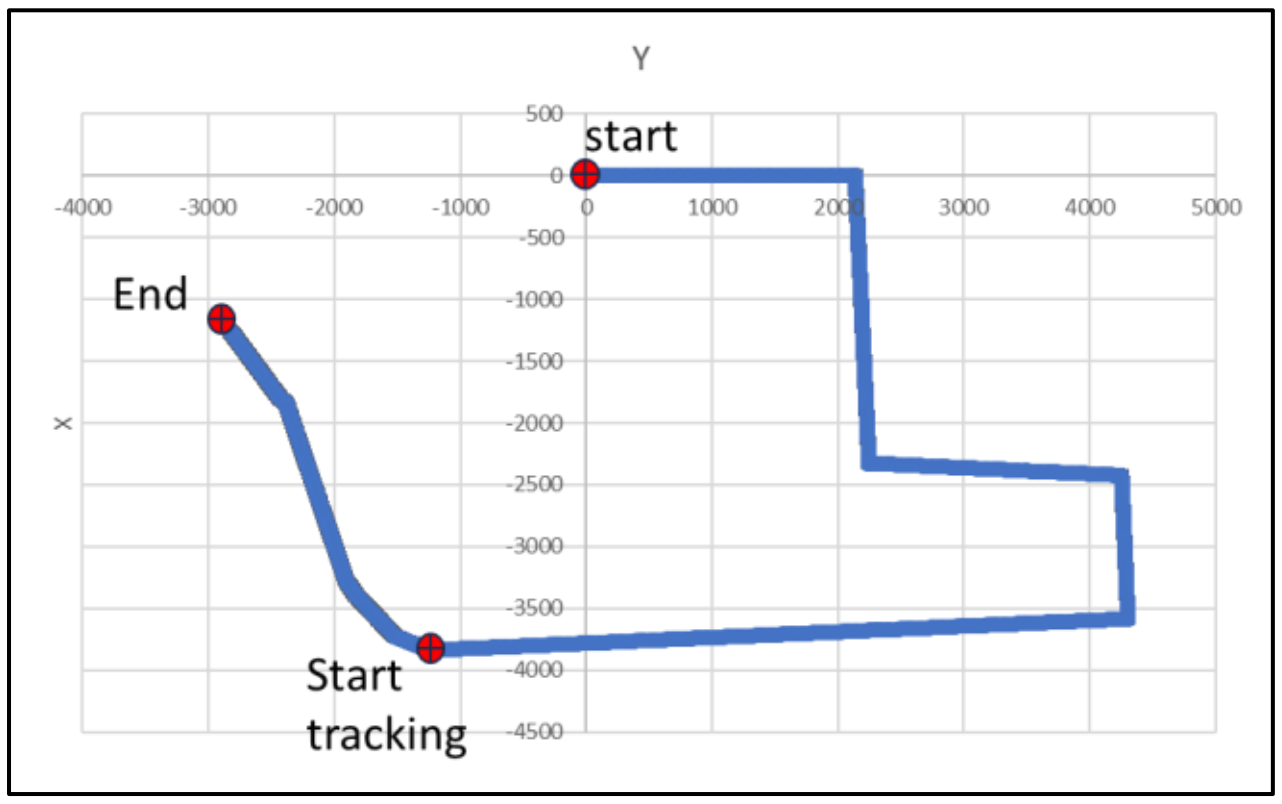

Figure 15. Dynamic path: Pioneer robot trajectory mapping from the first experiment.

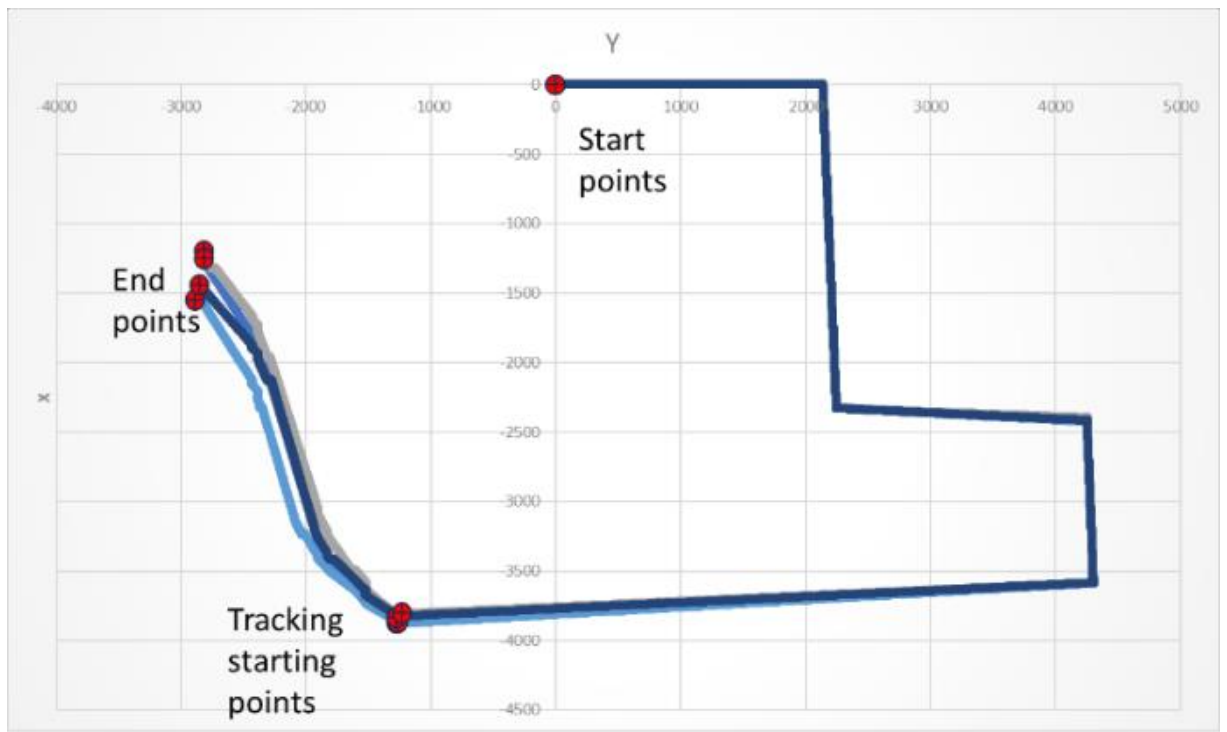

Figure 16. Overlay of multiple iterations of experimental trajectories from odometry and target tracking. (Here example shows 4 overlayed trajectories).

## V. Conclusions

This study has explored the integration of advanced navigation and control techniques in autonomous robot systems, focusing on odometry, target tracking, and obstacle avoidance. Through detailed method descriptions and experimental setups in simulated and real-world environments, the effectiveness of each behavior was demonstrated.

Odometry, utilizing wheel encoder data, enables precise navigation along predefined paths, validated through experiments with a Pioneer robot. Target tracking employs vision-based techniques for pursuing designated targets while avoiding obstacles, demonstrated on the same platform. Obstacle avoidance utilizes ultrasonic sensors to navigate cluttered environments safely, validated in both simulated and real-world scenarios.

Additionally, the study extended to include an Elegoo robot car, showcasing its features for enhanced experimentation. Through advanced algorithms and experimental validations, insights were provided into developing robust navigation systems for autonomous robots.

The contributions of this research extend beyond academia, offering practical solutions and insights applicable to various industries and domains. By equipping robots with advanced navigation and control capabilities, new possibilities emerge in fields such as autonomous transportation, industrial automation, and environmental monitoring. The development of a user control program facilitates remote operation and data exchange, enhancing the usability of robotic systems.

In conclusion, this study lays the groundwork for further advancements in robotics technology, forging the path for a future where autonomous robots play a crucial role in shaping society and advancing human progress.

## Acknowledgments

I express my heartfelt gratitude to Prof. Mary He and Dr. Salem Ameen for their steadfast support, which has profoundly influenced my academic journey. Their invaluable advice and guidance during this paper are deeply appreciated, as they have significantly enhanced the excellence of this work through their feedback and recommendations.

## References

[1] Thrun, S., Montemerlo, M., Dahlkamp, H., Stavens, D., Aron, A., Diebel, J., & Whittaker, W. (2005). "FastSLAM: An efficient solution to the simultaneous localization and mapping problem with unknown data association." In Proceedings of the Ninth International Conference on Artificial Intelligence and Statistics (pp. 598-605). IEEE.

[2] Milan, A., Rezatofighi, S. H., Shi, Q., Dick, A., & Reid, I. (2016). "Online multi-object tracking using CNN-based single object tracker with spatial-temporal attention mechanism." In Proceedings of the IEEE Conference on Computer Vision and Pattern Recognition (pp. 4293-4302).

[3] Liu, Z., Tang, L., & Chen, Z. (2018). "WiFi-based indoor localization using deep learning and signal strength measurements." In *Proceedings of the IEEE International Conference on Computer Communications* (pp. 1-6).

[4] Kadous, M. W., Nguyen, A., & Peters, R. A. (2019). "Cloud robotics: Concepts, technologies, and applications." *IEEE Transactions on Robotics*, 35(1), 276-291.

[5] Dissanayake, M. W., Newman, P., Clark, S., Durrant-Whyte, H. F., & Csorba, M. (2001). "A solution to the simultaneous localization and map building (SLAM) problem." In IEEE Transactions on Robotics and Automation, 17(3), 229-241.

[6] Thrun, S., Montemerlo, M., Dahlkamp, H., Stavens, D., Aron, A., Diebel, J., ... & Whittaker, W. (2005). "FastSLAM: An efficient solution to the simultaneous localization and mapping problem with unknown data association." In Proceedings of the Ninth International Conference on Artificial Intelligence and Statistics (pp. 598-605).

[7] Redmon, J., Divvala, S., Girshick, R., & Farhadi, A. (2016). "You only look once: Unified, real-time object detection." In Proceedings of the IEEE Conference on Computer Vision and Pattern Recognition (pp. 779-788).

[8] Liu, W., Anguelov, D., Erhan, D., Szegedy, C., Reed, S., Fu, C. Y., & Berg, A. C. (2016). "SSD: Single shot multibox detector." In European conference on computer vision (pp. 21-37).

[9] For the discussion on WiFi-based localization and cloud robotics: Gao, J., Chen, S., Tian, Y., Wang, W., Zhao, C., Zhao, Y., & Jin, Z. (2017). "A Wi-Fi-based indoor positioning system." In IEEE Communications Magazine, 55(1), 106-112.

[10] Kadous, M. W., Nguyen, A., & Peters, R. A. (2019). "Cloud robotics: Concepts, technologies, and applications." IEEE Transactions on Robotics, 35(1), 276-291.

[11] S. Thrun, W. Burgard, and D. Fox, Probabilistic Robotics. Cambridge, MA: MIT Press, 2005.

[12] R. Siegwart and I. R. Nourbakhsh, Introduction to Autonomous Mobile Robots. Cambridge, MA: MIT Press, 2004.

[13] S. M. LaValle, Planning Algorithms. Cambridge, U.K.: Cambridge University Press, 2006.

[14] J. Kang and H. Heo, "Navigation algorithm of mobile robot with omni-directional vision sensor," in Proc. Int. Conf. Control, Automation and Systems, 2006.

Lint to the source code, recorded result and video files:
Mobile_Robotice_HUSANVOKHIDOV.zip